# A Method for Extraction and Recognition of Isolated License Plate Characters

YON-PING CHEN
Dept. of Electrical and Control Engineering,
National Chiao-Tung University
Hsinchu, Taiwan
Email:ypchen@cc.nctu.edu.tw

TIEN-DER YEH
Dept. of Electrical and Control Engineering,
National Chiao-Tung University
Hsinchu, Taiwan
Email:tainder.ece91g@nctu.edu.tw

*Abstract*—A method to extract and recognize isolated characters in license plates is proposed. In extraction stage, the proposed method detects isolated characters by using Difference-of-Gaussian (DOG) function, The DOG function, similar to Laplacian of Gaussian function, was proven to produce the most stable image features compared to a range of other possible image functions. The candidate characters are extracted by doing connected component analysis on different scale DOG images. In recognition stage, a novel feature vector named accumulated gradient projection vector (AGPV) is used to compare the candidate character with the standard ones. The AGPV is calculated by first projecting pixels of similar gradient orientations onto specific axes, and then accumulates the projected gradient magnitudes by each axis. In the experiments, the AGPVs are proven to be invariant from image scaling and rotation, and robust to noise and illumination change.

*Keywords-accumulated gradient; gradient projection; isolated character; character extraction; character recognition*

## I. INTRODUCTION

The license plate recognition, or LPR in short, has been a popular research topic for several decades [1] [2] [3]. An LPR system is able to recognize vehicles automatically and therefore useful for many applications such as portal controlling, traffic monitoring, stolen car detection, and etc. Up to now, an LPR system still faces some problems concerning various light condition, image deformation, and processing time consumption [3].

Traditional methods for recognition of license plate characters often include several stages. Stage one is detection of possible areas where the license plates may exist. To detect license plates quickly and robustly is a big challenge since images may contain far more contents than just only expected ones. Stage two is segmentation, which divides the detected areas into several regions containing one character candidate. Stage three is normalization; some attributes of the character candidates, e.g., size or orientation, are converted to pre-defined values for later stages. Stage four is recognition stage; the segmented characters can be recognized by technologies such as vector quantization [4] or neural networks [5] [6]. Most works propose to recognize characters in binary forms so that they find thresholds [7] to depict the regions of interest in the detected areas. Normalization is not always necessary for all recognition methods. Some recognition methods need normalized characters so that they need more computations to normalize the character candidates before recognition. In this paper the detection stage and segmentation stage are merged into an extraction stage. And the normalization is not necessary because the characters are recognized in an orientation and size invariant manner.

The motivations of this work originate from three limitations of traditional method of LPR systems. First, traditional method uses simple features such as gradient energy to detect the possible location of license plates. However, this method may lose some plate candidates because the gradient energy may be suppressed due to camera saturation or underexposure, which often takes place under extreme light conditions such as sun light or shadow. Second, traditional detection methods often assume the license plates images are captured in a correct orientation so that the gradients can be accumulated on the pre-defined direction and then the license plates can be detected correctly. In real cases, the license plates may not always keep the same orientations in the captured images. They can be rotated or slanted due to the irregular roads, unfixed camera positions, or the abnormal conditions of cars. Third, it often happens that some characters in a license plate are blurred or corrupted which may fail the LPR process in detection or segmentation stage. The characteristic is dangerous for application because one single character may result in loss of whole license plate. Compare to human nature, people know the position of the unclear characters because they see some characters located aside. We try different methods, e.g. change head position or walk closer, to read the unclear character, and even guess it if it is still not distinguishable. This nature is not achievable in a traditional LPR system due to its coarse-to-fine architecture. To retain high detection rate of license plates under these limitations, the method in this paper proposes a fine-to-coarse method which firstly finds isolated characters in the captured image. Once some characters on a license plate are found, the entire license plate can be detected around these characters. The method may consume more computation than the traditional coarse-to-fine method. However, it minimizes the probability of missing license plate candidates in the detection stage.

A challenge to achieve the fine-to-coarse method is recognition of isolated characters. There are some difficulties to deal with isolated characters recognition. First, it is difficult

This research was supported by a grant provided by National Science Council, Taiwan, R.O.C.( NSC 98-2221-E-009 -128 -).

to extract orientation of an isolated character. In traditional LPRs [3], the orientations of characters can be determined by the baseline [3][8] of multiple characters. However this method is not suitable for isolated characters. Second, the unfixed camera view angle often introduces large deformation on the character shapes or stroke directions. It makes the detection and normalization process difficult to be applied. Third, the unknown orientations and shapes exposed under unknown light condition and environment become a bottleneck for the characters to be correctly extracted and recognized.

characters. Third, on each group the proposed accumulated gradient projection method is applied to find out the nature axes and associated accumulated gradient projection vectors (AGPVs). Finally, the AGPVs of each candidate are matched with those of standard characters to find the most similar standard character as recognition result. The experimental results show the feasibility of the proposed method and its robustness to several image parameters such as noise and illumination change.

## II. EXTRACTION OF ISOLATED CHARACTERS

Before extracting the isolated characters in an image, there are four assumptions made for the proposed methods:

1. The color (or intensities for gray scale images) of a character is monotonic, i.e., the character is composed of single color without texture on it.

2. Same as 1, the color of background around the character is monotonic, too.

3. The color of the character is always different from that of the background;

4. Characters must be isolated and no overlap in the input image.

In this chapter, the scale space theory is used and acts as the theoretical basis of the method to robustly extract the interested characters in the captured image. Based on the theory, the Difference-of-Gaussian images of all different scales are iteratively calculated and grouped to produce the candidates of license plate characters.

### A. Produce the Difference-of-Gaussian Images

Taking advantage of the scale-space theories [9]-[11], the extraction of characters becomes systematic and effective. In the first step, the detection of the characters is done by searching scale-space images of all the possible scales where the characters may appear in the input image. As suggested by the authors in [12] and [13], the scale-space images in this work are generated by convolving input image with different scale Gaussian functions. The first task to get the scale-space images is defining the Gaussian function. There are two parameters required for choosing of Gaussian filters, i.e., filter width $\lambda$ and smoothing factor $\sigma$, where the two parameters are not fully independent yet some relationship between them are required to be discussed.

The range of smoothing factor $\sigma$ is determined from experiments that a better choice of it is from 1 to 16 for the input image up to 3M pixels. There are two factors relevant to the sampling frequency of $\sigma$: the resolution of the target characters and the computational resources(including allowed processing time). These two factors play roles of trade-off and are often determined case by case. In this paper, we choose to set $\sigma$ of a scale double of that of the previous scale for convenient computation, i.e., $\sigma_2=2\sigma_1$, $\sigma_3=2\sigma_2$ ..., where $\sigma_1$, $\sigma_2$, $\sigma_3$..., are the corresponding smoothing factors of the scale numbered 1, 2, 3... As a result, the choice of smoothing factors in our case is, $\sigma_1=1$, $\sigma_2=2$, $\sigma_3=4$, $\sigma_4=8$, and $\sigma_5=16$. Consider

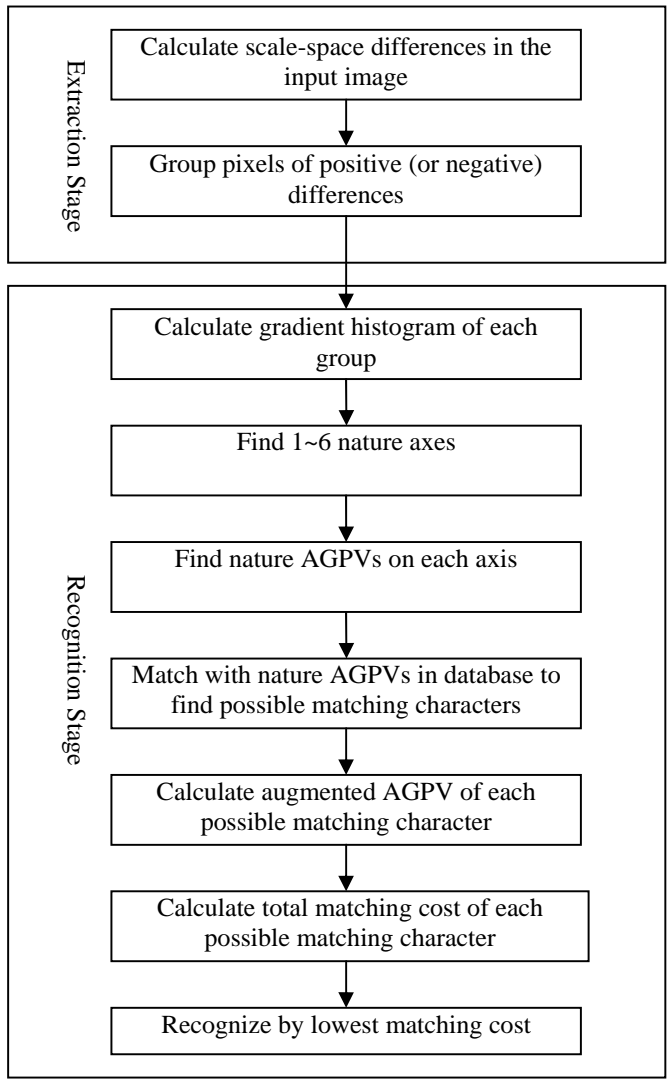

Figure 1. Process flow of the proposed method

The proposed scheme to extract and recognize license plate characters has procedures as the following. First, the candidates of characters are detected by the scale-space differences. Scale-space extrema has been proved stable against noise, illumination change and 3D view point change [9]-[14]. In this paper, the scale-space differences are approximated by the difference-of-Gaussian functions as in [9]. Second, the pixels of positive (or negative) differences are gathered into groups by connected components analysis and form candidates of factors of noise and sampling frequency in the spatial domain, a larger $\sigma$ is more stable to detect characters of larger sizes.

Ideally the width $\lambda$ of a Gaussian filter is infinity, while in real case it is reasonable to be an integer to match the design of digital filters. In addition, the integer cannot be large due to limited computation resources and only odd integers are chosen such that each output of convolution can be aligned to the center pixel of the filter. The width $\lambda$ is changed with the smoothing factor $\sigma$, which is in other words the standard deviation of the Gaussian distribution. Smaller $\sigma$ has better response on edges but yet more sensitive to noise. When $\sigma$ is small, there is no need to define a large $\lambda$ because the filter decays to a very small value when it reaches the boundary. In this paper we propose to choose the two parameters satisfying the following inequality,

$$\lambda \geq \sigma \times 7 \quad \& \quad \lambda = 2n+1, \forall n \in N. \quad (1)$$

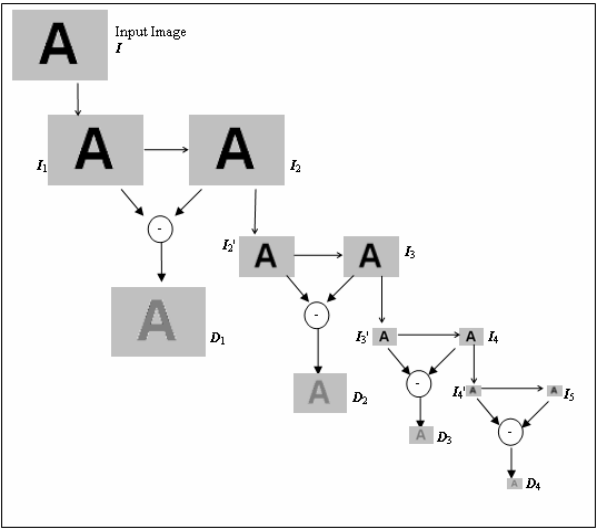

Figure 2. The procedure to produce difference-of-Gaussian Images

An efficient way to generate the smoothed images is taking sub-sampling. As explained above that the filter width is better chosen $\lambda \geq 7\sigma$, it makes the filter width grows to a large value when the smoothing factor grows up. This leads to a considerable amount of computation in real case if the filters are implemented in such a long size. To avoid expanding the filter width directly, we take use of sub-sampling on images of smoothing factors $\sigma>1$ based on the truth that the information in images is decreased as the smoothing factors increase.

Due to the fact that the image sizes vary with the level of sub-sampling, we store the smoothed images into a series of octaves according to their sizes. The images on an octave have one half length and width of those of the previous octave. On each octave there are two images subtracting from each other to produce the desired Difference-of-Gaussian (DOG) image for later processing. The procedure of producing the Difference-of-Gaussian images can be explained by Fig.2. Let the length and width of the input image $I(x,y)$ be $L$ and $W$ respectively. In the beginning, $I(x,y)$ is convolved with Gaussian filter $G(x,y,\sigma_a)$ to generate the first smoothed image, $I_1(x,y)$ for the first octave. $\sigma_a$ is the smoothing factor of the initial scale and is selected as 1 ($\sigma_a = \sigma_1$) in our experiments. The smoothed image $I_1(x,y)$ is used to convolve with Gaussian filter $G(x,y,\sigma_b)$ to generate the second smoothed image $I_2(x,y)$, which will be subtracted from $I_1(x,y)$ to generate the first DOG image $D_1(x,y)$ on the octave. The $I_2(x,y)$ is also sub-sampled by every two pixels on each row and column to produce the image $I_2'(x,y)$ for the next octave. It is worth to note that an image sub-sampled from a source image has smoothing factor equal to one half of that of the source image. The length and width of image $I_2'(x,y)$ are $L/2$ and $W/2$, and the equivalent smoothing factor is $(\sigma_a+\sigma_b)/2$ from initial scale. As the $\sigma_b$ is selected to be same as the smoothing factor $\sigma_a$ of the initial scale, the image $I_2'(x,y)$ therefore has the equivalent smoothing factor $\sigma=\sigma_a$, and is served as the initial scale of the second octave. The image $I_2'(x,y)$ is convolved with $G(x,y,\sigma_b)$ again to generate the third smoothed image, $I_3(x,y)$, which can be subtracted from $I_2'(x,y)$ to produce the second DOG image $D_2(x,y)$. The same procedure can be applied to the remaining octaves to generate the required smoothed images $I_4$ and $I_5$, and Difference-of-Gaussian images $D_3$ and $D_4$.

*B. Grouping of the Difference-of-Gaussian Images*

To find the interested characters in the DOG image, the next step is to apply connected components analysis to connect pixels of positive (or negative) responses into groups. After connected components analysis, all the groups are filtered by their sizes. There are expected sizes of characters for different octave and the groups will be discarded if their sizes are not falling into the expected range. The most stable sizes for extracting general characters on each octave are ranged from 32×32 to 64×64. Characters sizes smaller than 32×32 are easily disturbed by noise and result in undesirable outcomes. Characters sizes larger than 64×64 can be extracted on octaves of larger scales.

III. THE ACCUMULATED GRADIENT PROJECTION VECTOR(AGPV) METHOD

After extraction of the candidate characters, a novel method named accumulated gradient projection vector method, or AGPV method in short, is proposed and applied to recognize the extracted candidate characters. There are four stages to recognize a character using the AGPV method. First, determine the axes; including the nature axes and augmented axes. Second, calculate the AGPVs based on these axes. Third, normalize the AGPVs for comparing with standard ones. Fourth, match with standard AGPVs to validate the recognition result. The procedure will be explained in detail in the following sections.

*A. Determine Axes*

It is important to introduce the axes first before discussing the AGPV method. An axis of a character is a specific orientation on which the gradients of grouped pixels are projected and accumulated to form the desired feature vector.

An axis is represented by a line that has specific orientation and passes through the center of gravity point of the pixels group. The axes of a character can be separated into two different classes named nature axes and augmented axes, which are different in characteristics and usages and will be described below.

*1) Build up Orientation Histogram*

Upon an input image is clustered into one or more groups of pixels, the next step is to build up the corresponding orientation histograms. The orientation histograms are formed from gradient orientations of grouped pixels. Let $\gamma(x,y)$ be the intensity value of sample pixel $(x,y)$ of an image group $I$, the gradients on x-axis and y-axis are respectively,

$$\nabla X(x,y) = \gamma(x+1, y-1) - \gamma(x-1, y-1) + \\ 2 \times (\gamma(x+1, y) - \gamma(x-1, y)) + \gamma(x+1, y+1) - \gamma(x-1, y+1)$$
$$\nabla Y(x,y) = \gamma(x-1, y+1) - \gamma(x-1, y-1) + \\ 2 \times (\gamma(x, y+1) - \gamma(x, y-1)) + \gamma(x+1, y+1) - \gamma(x+1, y-1)$$
(2)

The gradient magnitude, $m(x,y)$, and orientation, $\theta(x,y)$, of this pixel is computed by

$$m(x,y) = \sqrt{(\nabla X(x,y))^2 + (\nabla Y(x,y))^2}$$
$$\theta(x,y) = tan^{-1}(\nabla Y(x,y) / \nabla X(x,y))$$
(3)

By assigning a resolution $BIN_{his}$ in the orientation histogram, the gradients are accumulated into $BIN_{his}$ bins and the angle resolution is $RES_{his} = (360/BIN_{his})$. The $BIN_{his}$ is chosen as 64 in the experiments and the angle resolution $RES_{his}$ is therefore 5.625 degrees. Each sample added to the histogram is weighted by its gradient magnitude and accumulated into the two nearest bins by linear interpolation. Besides the histogram accumulation, the gradient of each sample is accumulated into a variable $GE_{his}$ which stands for the total gradient energy of the histogram.

*2) Determine the Nature Axes*

The next step toward recognition is to find the corresponding nature axes based on the built orientation histogram. The word "nature" is used because the axes always exist "naturally" regardless of most environment and camera factors that degrade the recognition rate. The nature axes have several good properties helpful for the recognition. First, they have high gradient energy on specific orientation and therefore are easily detectable in the input image. Second, the angle differences among the nature axes are invariant to image scaling and rotation. It means, they can be used as references to correct the unknown rotation and scaling factors on the input image. Third, the directions of nature axes are robust within a range of focus and illumination differences. Fourth, although some factors, such as different camera view angle, may cause character deformation and change the angle relationship among the nature axes, the detected nature axes are still useful to filter out the dissimilar ones and narrow down the range of recognition.

Let function $H(a)$ denote the histogram magnitude appeared on angle $a$. Find the center of the $k$-th peak, $p_k$, of the histogram, which are defined by satisfying $H(p_k) > H(p_k -1)$ and $H(p_k) > H(p_k +1)$. A peak represents a specific orientation in the character image. Beside the center, find the boundaries of the peak, start angle $s_k$ and end angle $e_k$, within an threshold angle distance $a_{th}$, i.e.,

$$s_k = a, \ H(a) \leq H(b), \forall b \in (p_k - a_{th}, p_k)$$
(4)

$$e_k = a, \ H(a) \leq H(b), \forall b \in (p_k, p_k + a_{th})$$
(5)

The threshold $a_{th}$ is used to guarantee the boundaries of a peak stay nearby of its center and is defined to be 22.5 degrees in the experiment. The reason to choose ±22.5 degrees threshold is because it segments a 360-degree circle into 8 orientations, which is similar to human eyes often see a circle in 8 octants.

Once the start angle and end angle of a peak is determined, define the energy function of the $k$-th peak as $E(k) = \sum_{a=s_k}^{e_k} H(a)$, which stands for the gradient energy of a peak. In addition, an outstanding energy function $D(k)$ is also defined for each peak,

$$D(k) = E(k) - \frac{(H(s_k) + H(e_k)) \times (e_k - s_k)}{2}$$
(6)

The outstanding energy neglects the energy contributed by neighboring peaks and is more meaningful than $E(k)$ to represent the distinctiveness of a peak. Peaks with small outstanding energy are not considered as nature axes because they do not outstand from the neighboring peaks and may not be detectable in new images.

In the experiments, there are different strategies to threshold the outstanding energy for calculating standard AGPVs and test AGPVs. When calculating standard AGPVs, we select one grouped image as standard character image for each character and assign it to be the standard of the recognition. The mission of this task is to find stable peaks in the standard character image. Therefore, a higher threshold $GE_{his}/32$ is applied and a peak has outstanding energy higher than the threshold is considered as a nature axis of the standard character image. When calculating test AGPVs, the histogram may have many unexpected factors such as noise, focus error, bad lighting condition…, so that the task is changed to find one or more matched candidates for further recognition. Therefore, a lower threshold $GE_{his}/64$ is used to filter out the dissimilar ones by the outstanding energy. After threshold checking, the peaks whose outstanding energy higher than the threshold is called nature peaks of the character image and the corresponding angles are called the nature axes. Typical license plate characters can be found having two to six nature axes by the procedures above.

Fig.3 is an example to show the nature axes. Fig.3(a) is the source image, where intensity is ranged from 0(black) to 255(white). Fig.3(b) overlays the source image with the detected nature axes shown by red arrows. Fig.3(c) is the corresponding orientation histogram which are accumulated from the pixels gradient magnitudes in Fig.3(a). We can see six peaks in the histogram, marked as **A**,**B**,**C**,**D**,**E** and **F** respectively, which correspond to the six red arrows in Fig.3(b).

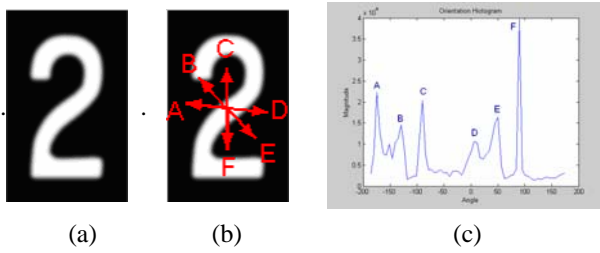

Figure 3. (a)input image (b)the nature axes (c)orientation histogram

*3) Determine the Augmented Axes*

Augmented axes are defined, as augmentations to nature axes, to be the directions on which the feature vectors, AGPVs, generated are unique or special to represent the source character. Unlike the nature axes possessing strong gradient energy on specific orientation, augmented axes do not have this property so that they may not be observed from orientation histogram.

Some characters, such as Fig.4, have only few (one or two) apparent nature axes. Therefore, it is necessary to generate enough AGPVs on augmented axes for the recognition process. The experiment tells us that a better choice for the number of AGPVs is four to six in order to recognize a character in a high successful rate. The AGPVs can be any one from nature axes AGPVs or augmented axes AGPVs.

The augmented axes can be defined by character shapes or by fixed directions. In our experiments, there are only four fixed directions, as the four arrows in Fig.4(b), defined as augmented axes for the total 36 characters. If any one of the four directions already exists in the nature axes, it will not be declared again in the augmented axes.

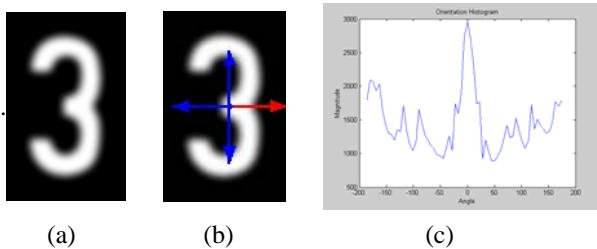

Figure 4. (a).a character has only one nature axis. (b)the nature axes in red arrow and three augmented axes in blue arrows(c)orientation histogram

*B. Calculate AGPVs*

Once the axes of a character are found, the next step is to calculate the accumulated gradient projection vectors (AGPVs) based on these axes. On each axis of corresponding peak $p_k$, the gradient magnitudes of pixels whose gradient orientations fall inside the range $s_k < \theta(x,y) < e_k$ are projected and accumulated. The axis could be any one in the nature axes or augmented axes.

*1) Projection principles*

The projection axis, $\eta_\phi$, is chosen from either nature axes or augmented axes with positive direction $\phi$. Let the ($x_{cog}$, $y_{cog}$) be the COG point of the input image, i.e.,

$$\begin{bmatrix} x_{cog} \\ y_{cog} \end{bmatrix} = \frac{1}{N} \times \begin{bmatrix} \sum_{i=1}^{N}(x_i) \\ \sum_{i=1}^{N}(y_i) \end{bmatrix}, \quad (7)$$

where ($x_i$, $y_i$) is the *i*-th sample pixel and *N* is the total number of sample pixels of a character. Let the function $A(x,y)$ denote the angle between sample pixel($x,y$) and the *x*-axis, i.e.,

$$A(x,y) = a\,tan\left(\frac{y}{x}\right). \quad (8)$$

The process of projecting a character onto axis $\eta_\phi$ can be decomposed into three operations. First, rotate the character by angle $\Delta\theta = (A(x_{cog}, y_{cog}) - \phi)$. Second, scale the rotated pixels by a projection factor $cos(\Delta\theta)$. And third, translate the axis origin to the desired coordinate. Apply the process on the COG point, the coordinate of COG point after rotation is,

$$\begin{bmatrix} x_{rcog} \\ y_{rcog} \end{bmatrix} = \begin{bmatrix} cos(\Delta\theta) & -sin(\Delta\theta) \\ sin(\Delta\theta) & cos(\Delta\theta) \end{bmatrix} \cdot \begin{bmatrix} x_{cog} \\ y_{cog} \end{bmatrix}, \quad (9)$$

Scaling by a projection factor $cos(\Delta\theta)$, it becomes,

$$\begin{bmatrix} x_{pcog} \\ y_{pcog} \end{bmatrix} = \begin{bmatrix} cos(\Delta\theta) & 0 \\ 0 & cos(\Delta\theta) \end{bmatrix} \cdot \begin{bmatrix} x_{rcog} \\ y_{rcog} \end{bmatrix}. \quad (10)$$

Finally, combine (15) and (16) and further translate the origin of axis $\eta_\phi$ to ($x_{\eta ori}$, $y_{\eta ori}$), the final coordinate ($x_{proj}$, $y_{proj}$) of projecting any sample pixel ($x,y$) onto axis $\eta_\phi$, is computed by

$$\begin{bmatrix} x_{proj} \\ y_{proj} \end{bmatrix} = \begin{bmatrix} cos^2(\Delta\theta) & -sin(\Delta\theta)cos(\Delta\theta) \\ sin(\Delta\theta)cos(\Delta\theta) & cos^2(\Delta\theta) \end{bmatrix} \cdot \begin{bmatrix} x \\ y \end{bmatrix} \quad (11)$$
$$- \begin{bmatrix} x_{pcog} \\ y_{pcog} \end{bmatrix} + \begin{bmatrix} x_{\eta ori} \\ y_{\eta ori} \end{bmatrix}$$

Note that the origin of axis $\eta_\phi$, $(x_{\eta ori}, y_{\eta ori})$, is chosen to be the COG point of the candidate character in the experiments, i.e., $(x_{\eta ori}, y_{\eta ori}) = (x_{cog}, y_{cog})$, because it concentrates the projected pixels around the origin $(x_{cog}, y_{cog})$ and saves some memory space used to accumulate the projected samples on new axis.

*2) Gradient projection accumulation*

In this section, the pre-computed gradient orientation and magnitude will be projected onto specific axes then summed up. Only sample pixels of similar gradient orientations are projected onto the same axis. See Fig. 5 for example; an object **O** is projected onto axis $\eta$ of angle 0-degree. In this case, only the sample pixels of gradient orientations $\theta(x,y)$ near 0-degree will be projected onto $\eta$ and then accumulated.

According to axes types, there are two different cases to select sample pixels of similar orientations. For nature axis corresponding to $k$-th peak $p_k$, the sample pixels with orientation $\theta(x,y)$ ranged inside the boundaries of the $p_k$, i.e., $s_k < \theta(x,y) < e_k$, are projected and accumulated. For augmented axis with angle $\phi$, the sample pixels with gradient orientations $\theta(x,y)$ ranged by $\theta(x,y) \geq \phi-22.5$ and $\theta(x,y) \leq \phi+22.5$ will be projected and accumulated.

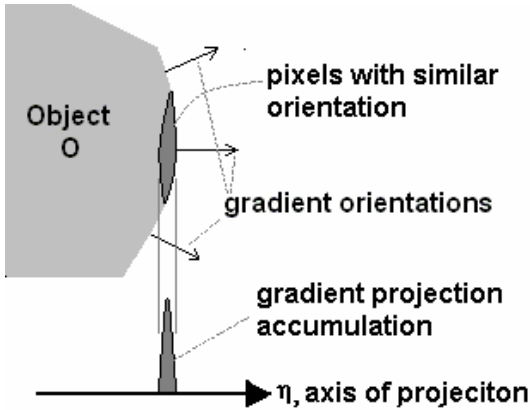

Figure 5.  Accumulation of gradient projection

From (3) and (11), the projected gradient magnitude, $\hat{m}(x,y)$, and the projected distance, $\hat{\ell}(x,y)$ of sample pixel $(x,y)$ onto axis $\eta_\phi$ are respectively

$$\hat{m}(x,y) = m(x,y) \times cos(\theta(x,y) - \phi) \quad (12)$$

$$\hat{\ell}(x,y) = \sqrt{(x_{proj} - x_{cog})^2 + (y_{proj} - y_{cog})^2} \quad (13)$$

To accumulate the gradient projections, an empty array $R(x)$ is created with length equals to the diagonal of the input image. Since the indexes of an array must be integers, linear interpolation is used to accumulate the gradient projections into the two nearest indexes of the array. In mathematical representations, let $b=floor(\hat{\ell}(x,y))$ and $u=b+1$, where $floor(z)$ rounds $z$ to the nearest integers towards minus infinity. For each sample pixel $(x,y)$ on input image $I$, do the following accumulations,

$$\begin{aligned} R(b) &= R(b) + \hat{m}(x,y) \times (\hat{\ell}(x,y) - b) \\ R(u) &= R(u) + \hat{m}(x,y) \times (u - \hat{\ell}(x,y)) \end{aligned} \quad (14)$$

Besides $R(x)$, an additional gradient accumulation array, $T(x)$ is also created to collect extra information required for normalization. There are two differences between $R(x)$ and $T(x)$. First, unlike $R(x)$ targeting on only the sample pixels of similar orientation, $T(x)$ targets on all the sample pixels of a character and accumulates their gradient magnitudes. Second, $R(x)$ accumulates the projected gradient magnitude $\hat{m}(x,y)$, while $T(x)$ accumulates the original gradient magnitude $m(x,y)$. Referring to (14),

$$\begin{aligned} T(b) &= T(b) + m(x,y) \times (\hat{\ell}(x,y) - b) \\ T(u) &= T(u) + m(x,y) \times (u - \hat{\ell}(x,y)) \end{aligned} \quad (15)$$

The purpose of $T(x)$ is to collect the overall gradient information of the interested character then applied to normalize array $R(x)$ into desired AGPV.

*3) Normalization*

The last step to find out the AGPV on an axis is to normalize the gradient projection accumulation array $R(x)$ into a fixed-length vector. With the fixed length, the AGPVs have uniform dimensionality and can be compared with standard AGPVs easily. Before the normalization, the length of AGPV, $L_{AGPV}$, has to be determined. Depends on the complexity of recognition targets, different length of AGPV may be selected to describe the distribution of projected gradients. In our experiments, the $L_{AGPV}$ is chosen as 32. A smaller $L_{AGPV}$ lowers the resolution and degrades the recognition rate. A larger $L_{AGPV}$ slows down system performance and makes no significant difference on recognition rate. It is worth to note that, one AGPV formed upon an axis is independent from the other AGPVs formed upon different axes. This is important to make the AGPVs independent from one another regardless of the formed character and axes.

In order to avoid the impact of isolated sample pixels which are mostly caused by noise, the array $R(x)$ is filtered by a Gaussian filter $G(x)$:

$$\tilde{R}(x) = R(x) * G(x), \quad (16)$$

where the operator * stands for convolution operation. The variance of the $G(x)$ is chosen as $\sigma = (Len_R)/128$ in the experiments, where $Len_R$ is the length of $R(x)$. It is found that this choice benefits in both resolution and noise rejection. Similarly, the array $T(x)$ is also filtered by the same Gaussian filter to eliminate the effect of noise. After Gaussian filtering, the array $T(x)$ is analyzed to find effective range, the range in which the data is effective to represent a character. The effective range starts from index $Xs$ and ends in index $Xe$, defined as

$$X_s = \{x_s, T(x_s) \geq th_T; T(x) < th_T, \forall x < x_s\}, \quad (17)$$

and

$$X_e = \{x_e, T(x_e) \geq th_T; T(x) < th_T, \forall x > x_e\}, \quad (18)$$

where the threshold $th_T$ is used to discard noise and is chosen as $th_T = Max(T(x))/32$ in the experiment. The effective range of $R(x)$ is the same as the effective range of $T(x)$, from $X_s$ to $X_e$.

As mentioned previously, the gradient projection accumulation results in a large sum along a straight edge. This is a good property if the interested character is composed of straight edges. However, some characters may consist of not only straight edges but also some curves and corners which only contribute small energy on array $R(x)$. In order to balance the contribution of different types of edges and avoid the disturbance from noise, a threshold is used to adjust the content of array $R(x)$ before normalization,

$$\hat{R}(x) = \begin{cases} 0, & \text{if } \tilde{R}(x) < th_R \\ 255, & \text{if } \tilde{R}(x) \geq th_R \end{cases}. \quad (19)$$

After finding the effective range and adjusting the content of array $R(x)$, the accumulated gradient projection vector (AGPV) is defined to resample from $\hat{R}(x)$,

$$AGPV(i) = \hat{R}\left(round\left(\left(\frac{i}{32}\right) \times (X_e - X_s) + X_s\right)\right). \quad (20)$$

Fig.6 gives an example of the gradient accumulation array $T(x)$, gradient projection accumulation array $R(x)$ and normalized AGPV. The example uses the same test image as Fig.3 and only one of the nature axes, axis E, is selected and printed. Similar to the method of finding the peaks of orientation histogram, the $k$-th peaks, $p_k$, on $R(x)$ is defined as $R(p_k) > R(p_k -1)$ and $R(p_k) > R(p_k +1)$. It can be observed that four peaks exist in Fig.6(c) and each of them represents an edge projected onto axis E.

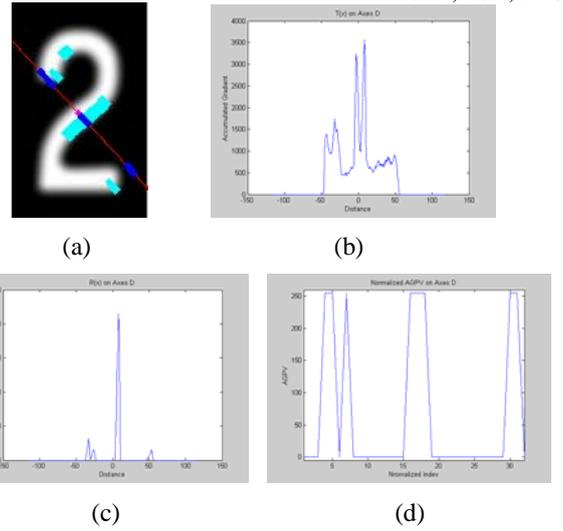

(a) (b)

(c) (d)

Figure 6. (a).Gradient projection on axis D. (pink: COG point; red: axis D; cyan: selected sample pixels; blue: locations after projection) (b).the gradient accumulation array $T(x)$ with distance to the COG point (c).the gradient projection array $R(x)$ (d).normalized AGPV.

### C. Matching and Recognition

#### 1) Properties used for matching

Unlike general vectors matching problem directly referring to the RMS error of two vectors, the matching of AGPVs refers to special properties derived from their physical meanings. There are three properties useful for similarity measuring between two AGPVs.

Each peak in an AGPV represents an edge on the source character. The number of peaks, or say the edge count, is useful to represent the difference between two AGPVs. For example, there are four peaks on the extracted AGPV in Fig.6(d) and each of them represents an edge on the axis. The edge count is invariant no matter how the character exists in the input image. In this paper, a function $EC(V)$ is defined to calculate the edge count of an AGPV $V$ and $EC(V)$ increase if $V(i)=0$ and $V(i+1) >0$.

Although the edge count in an AGPV is invariant for the same character, the position of the edges could be varied if the character is deformed due to slanted camera angle. This is the major reason to explain why the RMS error is not suitable to measure the similarity between two AGPVs. In order to compare AGPVs under the cases of character deformation, a matching cost function $C(U, V)$ is calculated to measure the similarity between AGPV $U$ and AGPV $V$, expressed as,

$$C(U,V) = |EC(U) - EC(V)| + |EC(UV) - EC(V)| + |EC(IV) - EC(V)| \quad , (21)$$

Where $UV = U \cup V$ is the union vector of AGPV $U$ and AGPV $V$ and $UV(i)=1$ if $V(i)>0$ or $U(i) >0$. $IV = U \cap V$ is the intersection vector and $IV(i)=1$ if $V(i)>0$ and $U(i) >0$.

An inter-axes property used to match the test character with the standard characters is that the angular relationships of nature axes on the test character are similar to those on the corresponding standard character. In the experiment, a threshold $th_A=\pi/32$ is used to check if the AGPVs of the test character match the angular relationship of nature axes of a standard character. Let $AA_T(k)$ be the $k$-th axis angle of the test character and $0 \le AA_T(k) < 2\pi$. The function $AA(i,j)$ denote the angle of the $j$-th axis of the $i$-th standard character, similarly $0 \le AA(i,j) < 2\pi$. If the $m$-th and $n$-th axis of the test character are respectively corresponding to the $g$-th and $h$-th axis of the $i$-th standard character, then

$$|(AA_T(m) - AA_T(n)) - (AA(i,g) - AA(i,h))| \le th_A. \quad (22)$$

*2) Create standard AGPV database*

A standard database is created by collecting all the AGPVs extracted from characters of standard parameters: standard size, standard aspect ratio, no noise, no blur, and neither rotation nor deformation. The extracted AGPVs are called standard AGPVs and stored by two categories: the one calculated on nature axes is called the standard nature AGPVs and the other calculated on augmented axes is called the standard augmented AGPVs. Let the number of total standard characters be $N$, $N=36$ (0~9 and A~Z) for license plate characters in this paper. Denote the number of standard nature AGPVs for $i$-th standard character as $NN(i)$, the number of standard augmented AGPVs as $NA(i)$, and the total number of AGPVs as $NV(i)$, where $NV(i)= NN(i)+NA(i)$. The $j$-th standard AGPV of the $i$-th character is denoted as $V_S(i,j)$, where $j=1$ to $NV(i)$. Note that $V_S(i,j)$ are standard nature AGPVs for $j \le NN(i)$ while $V_S(i,j)$ are standard augmented AGPVs otherwise.

*3) Matching of characters*

In order to recognize the test character, the AGPVs of it is stage-by-stage compared with the standard AGPVs in the database. Moreover, a candidates list is created by including all the standard characters at the beginning and removes those having high matching cost to the test character on each stage. Until the end of the last stage, the candidate in the list consisting of the lowest total matching cost is considered as the recognition result.

Stage 1: Find the fundamental matching pair. Calculate the cost function between the test character and the $j$-th AGPV of the $i$-th standard character.

$$C_1(k,j) = MC(V_T(k), V_S(i,j)) \quad (23)$$

Find a pair of axes whose matching cost is minimum:

$$(k_T, j_S) = \arg\min_{k,j}(C_1(k,j)) \quad (24)$$

If $min(C_1(k_T, j_S))$ is less than a threshold $th_F$, the $i$-th standard character is kept in the candidates list and the pair $(k_T, j_S)$ is served as the fundamental pair of the candidate.

Stage 2: Find the other matching pairs between the standard AGPVs and the test character: Based on the fundamental pair, the axes angles of the test character are compared with those of the standard character. Let the number of nature AGPVs detected on the test character be $NN_T$. For the $i$-th standard character, create an empty array $mp(j)=0$, $1 \le j \le NV(i)$, to denote the matching pair with the test character. Take use of (22), calculate

$$|(AA_T(k) - AA_T(k_T)) - (AA(i,j) - AA(i,j_S))| \le th_A$$
$$\forall k \in [1, NN_T], k \ne k_T; \forall j \in [1, NN(i)], j \ne j_S \quad (25)$$

the $k$-th test AGPV satisfies (25) is called the $j$-th matching pair of the standard character, denoted as $mp(j)=k$. Note that there might be more than one test AGPVs satisfying (25). In this case only the one of lowest matching cost is recognized as the $j$-th matching axis and the others are ignored.

Stage 3: Calculate total matching cost of standard nature AGPVs: Define a character matching cost function $CMC(i)$ to measure the similarity between test character and the $i$-th standard character by summing up the matching costs of all the matching pairs,

$$CMC(i) = \sum_{j=1, mp(j)>0}^{NN(i)} MC(V_T(mp(j)), V_S(i,j)) \quad (26)$$

Stage 4: Calculate the matching costs of augmented AGPVs: At the first step, find the axis angle $AX$ on the test character corresponding to the $j$-th standard augmented axis as

$$AX = (AA(i,j) - AA(i,j_S)) + AA_T(k_T) \quad (27)$$

If there is one AGPV of the test character, say, the $k$-th nature AGPV, satisfies (25), i.e., $|AA_T(k) - AX| \le th_A$, then the $k$-th nature AGPV is mapped to the $j$-th augmented axis and $mp(j)=k$. Otherwise, the AGPV corresponding to the $j$-th standard augmented axis must be calculated based on the axis angle $AX$. After that, the matching costs of the augmented AGPVs are accumulated into the character matching cost function as,

$$CMC(i) = CMC(i) + \sum_{j=NN(i)+1}^{NV(i)} MC(V_T(mp(j)), V_S(i,j)) \quad (28)$$

Stage 5: Recognition: Due to the different number of AGPVs for different standard character, the character matching cost function is normalized by the total number of standard AGPVs, i.e.,

$$CMC(i) = CMC(i) / NV(i) \quad (29)$$

Finally, the test character is recognized as the *h*-th standard character of the lowest matching cost if the character matching cost $CMC(h) < th_R$

## IV. EXPERIMENTAL RESULTS

The test images are captured from the license plates under general conditions and include several factors which often degrade recognition rates, such as dirty plates, deformed license plates, plates under special light conditions …, etc. All the images are sized 2048x1536 and converted into 8-bit gray-scale images. Total 60 test images are collected and each of them contains one license plates consisting of 4 to 6 characters. The minimum character size in the image is 32×32 pixels. We choose some characters from the test images to be the standard characters and calculate the standard AGPVs.

The results are measured by the true positive rate(TPR), i.e., the rate that the true characters are extracted and recognized successfully, and the false positive rate(FPR), i.e., the rate that the false characters are extracted and recognized as a character in the test image. The process is divided into two stages, extraction stage and recognition stage, and the result of each stage is recorded and listed in table I.

The discussion is focused on the stability of the proposed method with respect to the three imaging parameters: noise, character deformation, and illumination change. These parameters are considered to be the most serious factors to impact recognition rate. Denote the original test images as set A. Three extra image sets, set B, set C and set D, are generated to test the impact of these parameters, respectively.

In Table I, the column "Extraction TPR" stands for the rate that the characters in test images are extracted correctly. A character is considered as successfully extracted if, first, it is isolated from external objects and second, the grouped pixels can be recognized by human eyes. The column "Recognition 1 TPR" represents the rate that the extracted true characters are recognized correctly, under the condition that the character orientation is unknown. Nevertheless, the column "Recognition 2 TPR" is the recognition rate based on the condition that the character orientation is known so that the fundamental pair of a test character is fixed. This is reasonable for most cases since the license plates are often orientated horizontally. Under such conditions, the characters are kept in vertical orientations if the camera capture angle keeps upright.

TABLE I. SIMULATION RESULTS OF THE FOUR IMAGE SETS

|  | Extraction | Recognition 1 Unknown orientation | | Recognition 2 Known orientation | |
|---|---|---|---|---|---|
|  | TPR | TPR | FPR | TPR | FPR |
| Set A | 93.3 | 88.3 | 8.3 | 93.3 | 3.3 |
| Set B | 83.0 | 85.4 | 8.1 | 88.3 | 5.0 |
| Set C | 91.6 | 67.3 | 13.3 | 75.8 | 10.6 |
| Set D | 89.7 | 78.4 | 8.6 | 89.3 | 3.3 |

### A. Stability to Noise

The image set B is generated by two steps: First, copy set A to a new image set. Second, add 4% pepper and salt noise onto the new image set to form set B. Note that 4% pepper and salt noise means one pepper or salt appearing in every 25 pixels.

It can be seen from Table I that the character extraction rate is degraded when noise is added. A further experiment shows that enlarging the size of characters is very useful to improve the extraction rate under the effect of noise. It is reasonable since the noise energy is lowered if the range of accumulation is enlarged. Similarly, the same method is able to improve the TPR of recognition since it accumulates the gradient energy before feature extraction.

### B. Stability to Character Deformation

The image set C is generated by transforming pixels of set A via affine transformation matrix *M*, expressed as

$$\begin{bmatrix} x_C \\ y_C \end{bmatrix} = M \cdot \begin{bmatrix} x_A \\ y_A \end{bmatrix}. \quad (30)$$

Table II gives two examples of different matrix *M* and corresponding affine-transformed characters

TABLE II. TWO OF THE AFFINE-TRANSFORMED CHARACTERS

| Input \ *M* | $\begin{bmatrix} 1 & 0 \\ -1/6 & 1 \end{bmatrix}$ | $\begin{bmatrix} 1 & 0 \\ -1/4 & 1 \end{bmatrix}$ | $\begin{bmatrix} 1 & 0 \\ 1/6 & 1 \end{bmatrix}$ | $\begin{bmatrix} 1 & 0 \\ 1/4 & 1 \end{bmatrix}$ |
|---|---|---|---|---|
| A | A | A | A | A |
| E | E | E | E | E |

In Table I, the extraction TPR for set C is very close to the original rate for set A. This is because the extraction method by scale-space difference is independent from character deformation. However, the character deformation has serious impact to recognition TPR because not only the angle differences of the axes but also the peaks in AGPV are changed seriously due to the deformation.

A method to increase the recognition TPR for deformed input characters is expanding the database by including the deformed characters into standard characters. In other words, the false recognition can be reduced by considering the seriously deformed characters as new characters, then recognize it based on the new standard AGPVs. This method is helpful to resolve the problem of character deformation but takes more time in recognition as the AGPVs in standard database grow up comparing to the original.

### C. Stability to Illumination Change

It is found that the successful rate is robust and almost no change on extraction stage when the illumination is changed by constant factors, i.e.,

$$I'(x,y) = k \cdot I(x,y) \quad (31)$$

Consider to uneven illumination, four directional light sources, $L_1$ to $L_4$, are added into the test images and form Set D to imitate the response of illumination change. Expressed as

$$L_1(x,y) = (x+y+10)/(L+W+10)$$
$$L_2(x,y) = ((W-x)+y+10)/(L+W+10)$$
$$L_3(x,y) = (x+(L-y)+10)/(L+W+10)$$
$$L_4(x,y) = ((W-x)+(L-y)+10)/(L+W+10), \quad (32)$$

where the $W$ and $L$ are respectively the width and length of the test image. We can see from Table I that the uneven illumination change makes little difference to character extraction. A detail analysis indicates that insufficient color (intensity) depth makes some edges disappeared under illumination change and forms the major reason for the drop of extraction TPR. Similarly, the same reason also degrades the TPR in recognition stage.

A good approach to minimize the sensitivity to illumination change is to increase the color (intensity) depth of the input image. 12-bit or 16-bit gray-level test images will have better recognition rates than 8-bit ones.

## V. CONCLUSION

A method to extract and recognize the license plate characters is presented comprising two stages: First, extract isolated characters in a license plate. And second, recognize them by the novel AGPVs.

The method in extraction stage incrementally convolves the input image with different scale Gaussian functions and minimizes the computations in high scale images by means of sub-sampling. The isolated characters are found by connected components analysis on the Difference-of-Gaussian image and filtered by expected sizes. The method in recognition stage adopts AGPV as feature vector. The AGPVs calculated from Gaussian-filtered images are independent from rotation and scaling, and robust to noise and illumination change.

The proposed method has two distinctions with traditional methods:

1. Unlike traditional methods detect the whole license plate in the first step; the method proposed here extracts characters alone and no need to detect the whole plate in advance.

2. The recognition method is suitable for single characters. Unlike traditional methods require the information of baseline before recognition; the proposed method requires no prior information of character sizes and orientations.

A direction for future research can be categorized to extend the method to recognize different font types. Not only license plate characters, a lot of isolated characters are wide spreading around our daily lives. For example, there are many isolated characters exist in elevators, clocks, telephones …, etc. Traditional coarse-to-fine methods to recognize license plate characters are not suitable for these cases because each of them have different background and irregular characters placement. The proposed AGPV method shall be useful to recognize these isolated characters if it can be adapted to different font types.